\documentclass[conference]{IEEEtran}
\IEEEoverridecommandlockouts
\usepackage{cite}
\usepackage{amsmath,amssymb,amsfonts}
\usepackage{algorithmic}
\usepackage{graphicx}
\usepackage{textcomp}
\usepackage{xcolor}
\def\BibTeX{{\rm B\kern-.05em{\sc i\kern-.025em b}\kern-.08em
    T\kern-.1667em\lower.7ex\hbox{E}\kern-.125emX}}

\usepackage{lipsum}
\usepackage{multirow}
\usepackage{subcaption}
\usepackage[export]{adjustbox}
\usepackage{hyperref}
\usepackage[inline]{enumitem}
\usepackage{balance}
\usepackage[export]{adjustbox}
\begin{document}
\bstctlcite{IEEEexample:BSTcontrol}

\title{Productive Reproducible Workflows for DNNs: \\ A Case Study for Industrial Defect Detection}


\author{Perry Gibson, Jos\'e Cano \\
\emph{School of Computing Science, University of Glasgow, UK} \\
}

\maketitle

\begin{abstract}

As Deep Neural Networks (DNNs) have become an increasingly ubiquitous workload, the range of libraries and tooling available to aid in their development and deployment has grown significantly.
Scalable, production quality tools are freely available under permissive licenses, and are accessible enough to enable even small teams to be very productive.
However within the research community, awareness and usage of said tools is not necessarily widespread, and researchers may be missing out on potential productivity gains from exploiting the latest tools and workflows.
This paper presents a case study where we discuss our recent experience producing an end-to-end artificial intelligence application for industrial defect detection.
We detail the high level deep learning libraries, containerized workflows, continuous integration/deployment pipelines, and open source code templates we leveraged to produce a competitive result, matching the performance of other ranked solutions to our three target datasets.
We highlight the value that exploiting such systems can bring, even for research, and detail our solution and present our best results in terms of accuracy and inference time on a server class GPU, as well as inference times on a server class CPU, and a Raspberry Pi 4.

\end{abstract}

\begin{IEEEkeywords}
deep learning, docker, defect detection, pytorch, reproducibility, bonseyes
\end{IEEEkeywords}

\section{Introduction}
\label{sec:intro}

\begin{figure}
\centering
    \includegraphics[width=0.54\linewidth]{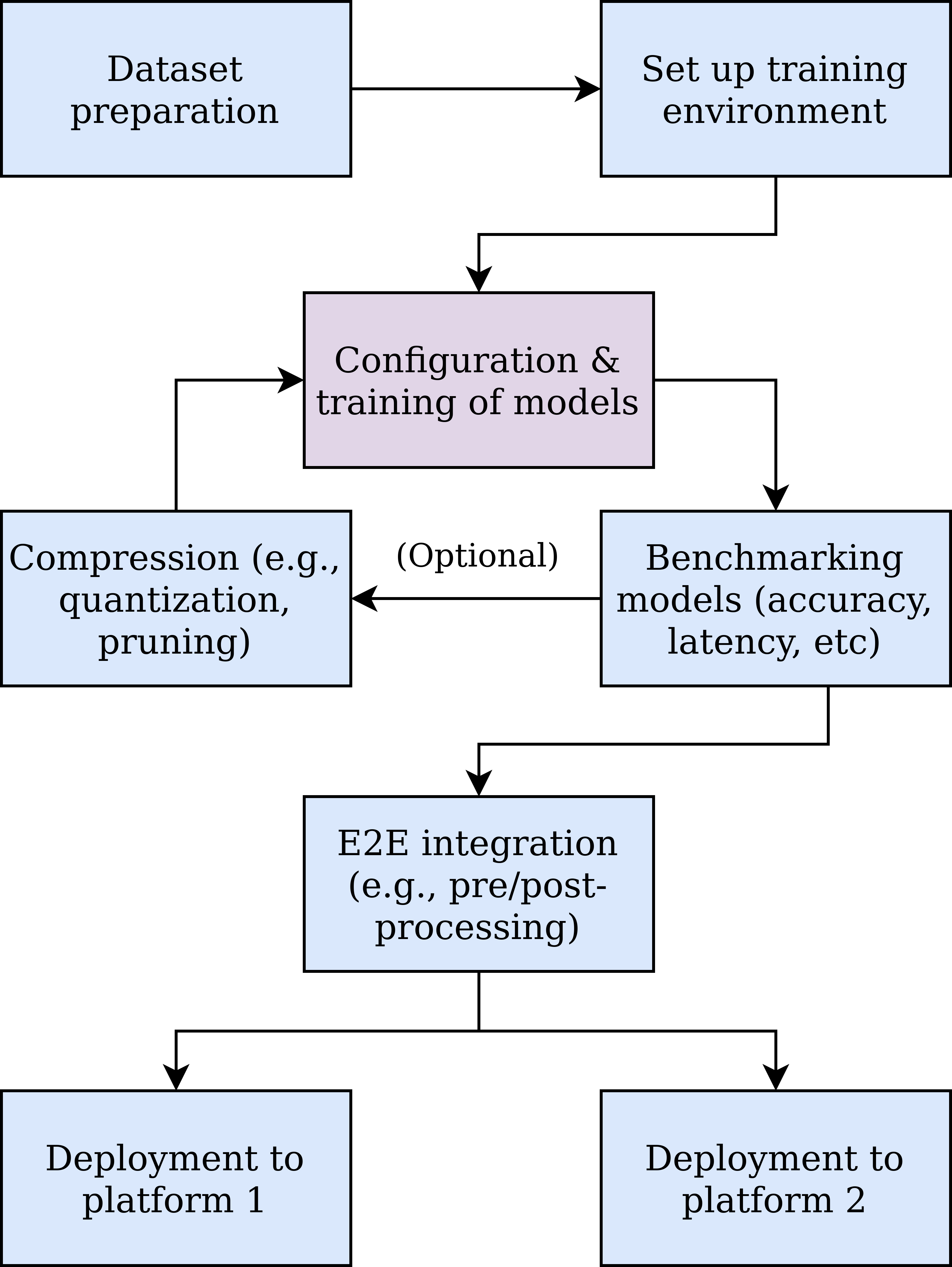}
\caption{
Simplified representation of the workflow of developing a DNN application.
Note that only one step directly involves choosing model architectures and training.
}
\label{fig:dnn_workflow}
\end{figure}

Deep Learning is becoming a common component within applications for a number of domains, from computer vision~\cite{he2016deep,denseNet2017,long2015Segmentation,yolo2016}, natural language processing~\cite{cnnsentence2014,Kalchbrenner14aconvolutional}, scientific computing~\cite{troster2019painting,kutz2017,mater2019}, and many more.
To aid in the development of these applications, there are a wide range of libraries and tools, such as deep learning frameworks including PyTorch~\cite{paszke2017automatic}, TensorFlow~\cite{abadi2016}, and MXNet~\cite{chen2015mxnet}.
However, beyond the creation of models themselves, there are a number of supplementary steps for creating applications that leverage Deep Neural Networks (DNNs).
As shown in Figure~\ref{fig:dnn_workflow}, many of the steps involved in developing a DNN application do not directly involve DNN models (e.g., dataset preparation, setting up development and deployment environments).
Researchers and industry practitioners generally take care when designing (or choosing) and training their models.
However, many of these complementary steps may be overlooked or implemented in a more ad-hoc manner, especially within the research community.
However, we argue that there are advantages to leveraging the growing set of tools and workflows for end-to-end DNN application development in research, even if the end goal is not production ready deployment.
For instance, ensuring that datasets are in a consistent and easily usable format for DNN training, and that this transformation process is reproducible.
Or when evaluating on more than one hardware platform, ensuring that the software environment is correctly set up with all of the required software dependencies, and ideally in a way which is reproducible.
For the latter example, continuous integration (CI) and continuous deployment (CD) pipelines can fit this role, however can be time consuming and tedious to set up from scratch.
Thus, in this position paper we discuss key tools which increased our productivity in developing an artificial intelligence (AI) application for visual industrial defect detection, and how integrating such tools into DNN development workflows can help both researchers and industry practitioners. 

The contributions of this paper include the following:

\begin{itemize}
    \item We describe several tools which we have used to increase the productivity of our DNN research, including PyTorch Lightning~\cite{Falcon_PyTorch_Lightning_2019} and templates from the Bonseyes Marketplace Platform~\cite{llewellynn2017}.
    
    \item We highlight how these tools were valuable to us in a case study for visual industrial defect detection, and how we used them to develop an end-to-end solution.
    
    \item We describe the three datasets we used to tackle our problem, and the models we trained.
    We present our best performing models in terms of accuracy and inference time, using an Nvidia A100 as our main evaluation platform, as well as presenting results on an x86 CPU, and a Rapsberry Pi 4.
\end{itemize}

\section{Core DNN development/deployment tools}
\label{sec:core_tools}

Achieving high accuracy on a target problem is generally the main motivating goal of any machine learning project, while latency becomes more important in systems research and when deploying to constrained devices in industrial use-cases.
Machine learning researchers can explore a wide range of design choices, for example varying the neural architecture, changing aspects of the training process (e.g., learning rate, optimizer), and applying varying types of data augmentation.

However, although these aspects of the solution are pivotal, it is important that the supporting infrastructure to help solve the problem is not overlooked or chosen as an afterthought.
For example: how is the raw training data to be translated into a format that the DNN can understand?
Can this be easily reproduced?
What is the software environment that a DNN will be trained in, and will it still work in future when packages are updated?
What platforms will the DNN be deployed to, and how will this deployment be managed?
These questions are important, thus in this paper we list a number of open source tools that we leveraged for our case study (discussed in Section~\ref{sec:case_study}), and the value they can bring for deep learning application development.
We do not list the most obvious and ubiquitous tools, for example PyTorch~\cite{paszke2017automatic}, which is the most popular deep learning framework used in research~\cite{paperswithcode2022}, or version control systems such as git.
Instead we focus on systems and tools which we believe filled a niche that greatly increased our productivity in carrying out our case study discussed in Section~\ref{sec:case_study}, and may not necessarily be well known or commonly used within the research community.
In particular, instrumental to our success were systems and templates provided by the Bonseyes Marketplace~\cite{llewellynn2017}, which were designed with these goals in mind~\footnote{More information on the Bonseyes suite of tools can be found in the Bonseyes Platform documentation~\cite{bonseyes2022}.}.
The core tools we leveraged were as follows:

\textbf{Segmentation Models PyTorch (SMP)}~\cite{Yakubovskiy:2019}: a library which builds on top of PyTorch, and eases the development of DNN applications for computer vision problems related to image segmentation.
Since our case study is for industrial visual defect detection, which is a sub-problem of image segmentation, exploiting this library enabled us to produce a range of solutions more quickly than if we had created our own solution from scratch.
Frameworks and libraries for specific problem spaces, which build on top of lower level DNN frameworks (e.g., PyTorch and TensorFlow~\cite{abadi2016}) are becoming more popular, and researchers should be aware of them when approaching a new problem domain, since they may provide shortcuts to a solution, or at least provide a convenient set of benchmarks to compare against.
As well as SMP for image segmentation, other examples of higher level DNN libraries include the TensorFlow Object Detection API~\cite{huang2017_tfapi} for object detection, and HuggingFace's Transformers library~\cite{wolf2020} for natural language processing and other tasks suited for Transformer-based~\cite{vaswani2017attention} architectures.

\begin{figure}[th]
\centering
    \includegraphics[width=0.98\linewidth]{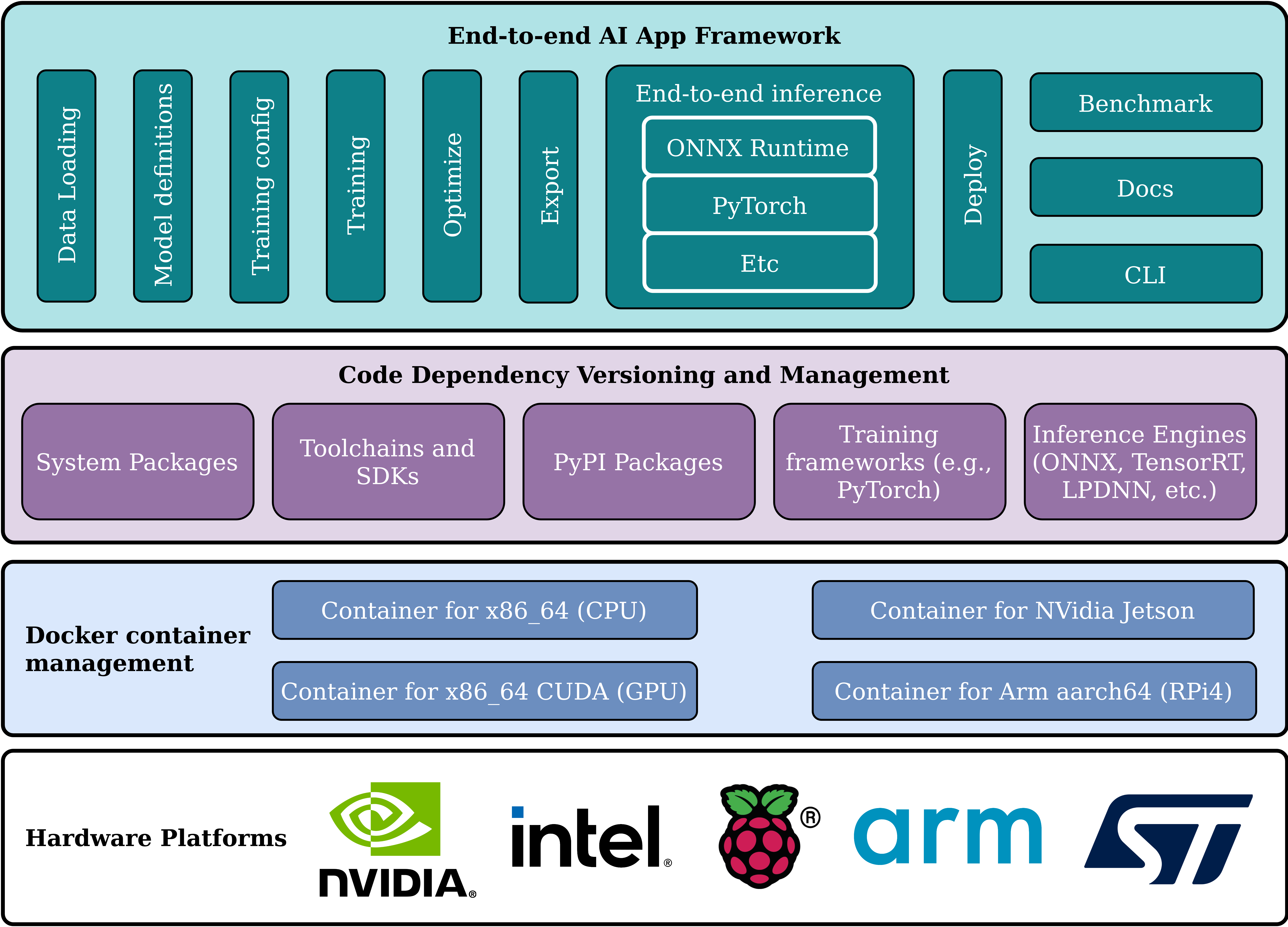}
\caption{
Simplified representation of the features provided by the Bonseyes AI Asset template system, adapted (with permission) from the AI Asset Generator documentation~\cite{bonseyesassociation2022}.
}
\label{fig:ai_asset}
\end{figure}

\textbf{PyTorch Lightning}~\cite{Falcon_PyTorch_Lightning_2019}: a wrapper library for PyTorch which reduces the amount of boilerplate code for defining model training.
In addition, it also provides utilities for pruning and quantization, which are designed to be simpler to use than those provided by normal PyTorch alone.
It also boasts the feature of enabling training across on multiple-GPUs, TPUs (Tensor Processing Units), CPUs, and IPUs (Intelligence Processing Units) without requiring changes in the code.

\textbf{Bonseyes Datatools}: a codebase template that allows developers to produce tools which convert raw data into a user-defined standard dataset format, including utilities for exploratory data analysis, visualization, and dataset tagging and versioning.
As well as ensuring that dataset preparation is more reproducible, a secondary purpose of a datatool is to separate initial dataset preparation from the model training code, which aids re-usability for future projects.

\textbf{Bonseyes AI Assets}: a codebase template that aids developers in producing a tool for both training and deployment of DNNs.
An AI Asset encapsulates all code and dependencies required for their solution, with an overview of its features shown in Figure~\ref{fig:ai_asset}.
Datasets generated from user-defined Bonseyes datatools can be easily mounted on the AI Asset, simplifying the data loading process.
Code for common activities such as benchmarking, report generation, and model conversion and inference using PyTorch, ONNX Runtime~\cite{onnxruntime}, and TensorRT~\cite{nvidiacorporation2016tensorrt} is provided, with support for more inference engines in development.
The design philosophy is to provide as much boilerplate code as possible without forcing developers to make design choices they do not want to.
Developers can use their deep learning framework of choice, and include any software dependencies they require.
The motivation for having all of the tools in a single environment is so that it is easier to investigate performance degradation over the whole pipeline, even when deploying on other platforms.
The trade-off here is increased disk storage for libraries.

\textbf{Bonseyes AI Asset CI/CD pipeline}: Continuous Integration (CI) and Continuous Deployment (CD) are software engineering principles whereby code is regularly subjected to automated testing, with CD being particularly focused on ensuring that code works in a deployment environment.
Although valuable, setting up these pipelines can be a very time consuming task and may not be a high priority for researchers who are focused on validating their ideas rather than producing production ready systems.
However, the Bonseyes AI Asset includes a predefined CI/CD pipeline, which means that developers can reap the benefits of having their development and deployment environments be independently tested with each code commit without requiring the high initial set-up costs.
Users must provide an x86-based server featuring an NVidia GPU and run a setup script which allows the server to receive and test new code commits, automatically testing for four platforms (x86+CUDA, x86-only, Nvidia Jetson, and Raspberry Pi, as seen in Figure~\ref{fig:ai_asset}).
QEMU~\cite{bellard2005qemu} emulation is used to test the Arm-based Jetson and Raspberry Pi platforms on the server, with Docker containers for each platform being generated and available for immediate deployment at the end of the process.
The testing process is automatic, with developers being sent an email if their pipeline fails.


\section{Case Study}
\label{sec:case_study}

As discussed in Section~\ref{sec:intro}, our goal was to produce an end-to-end AI application for the problem of industrial visual defect detection.
In essence, the task is to take visual input (e.g., from a camera) of some industrial product (e.g., textiles, rolled steel, printed circuit boards, etc) and identify if there are any defects on the product (e.g., scratches, blemishes, smudges, etc).
This information can then be used to improve product quality, and reduce waste.
An example of this can be seen in Figure~\ref{fig:sample_defect}, where in Figure~\ref{fig:first} we can see a photograph from some industrial product, and in Figure~\ref{fig:sample_label} we have a human annotated label of where in the image a defect is, shown in red.
To solve this problem effectively we were required to have our data in a consistent format (Section~\ref{subsec:datatools}), have models which can efficiently process said data (Section~\ref{subsec:models}), and have other parts of our development and deployment workflow be as supportive as possible for our workflow (Section~\ref{subsec:workflow}).


\subsection{Datasets and data processing}
\label{subsec:datatools}

\begin{figure}
\centering
\begin{subfigure}{0.49\linewidth}
\centering
    \includegraphics[height=0.8\linewidth, angle=90]{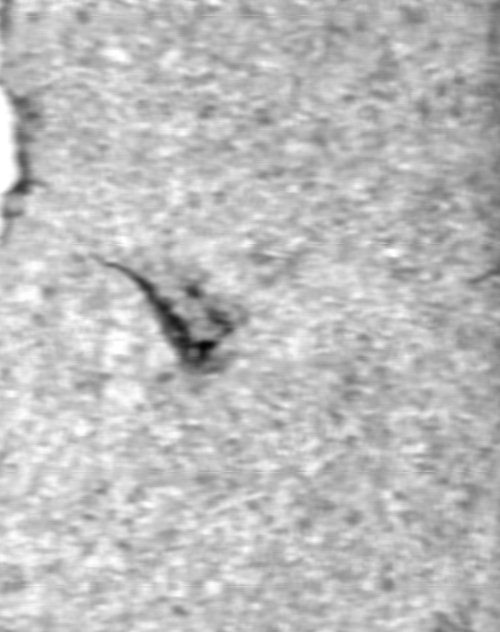}
    \caption{
    Image data
    }
    \label{fig:first}
\end{subfigure}
\hfill
\begin{subfigure}{0.49\linewidth}
\centering
    \includegraphics[height=0.8\linewidth, angle=90, frame]{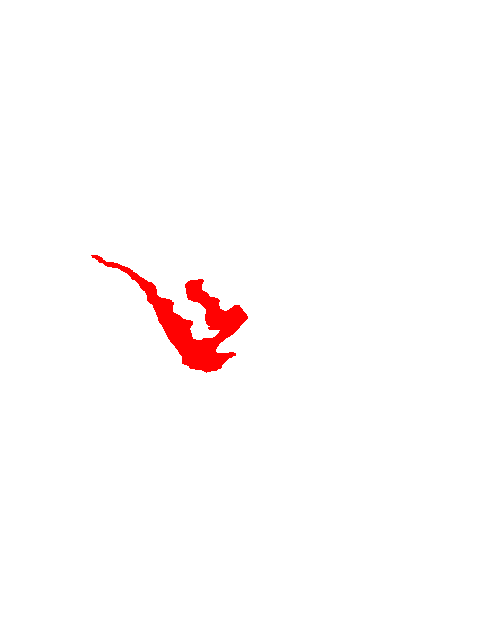}
    \caption{
    Defect label
    \label{fig:sample_label}}
\end{subfigure}
\caption{Sample element from the KolektorSDD dataset~\cite{Tabernik2019JIMkolek}, with the image data (a), and pixel-wise mask of defective regions highlighted in red (b).}
\label{fig:sample_defect}
\end{figure}

For our case study, we used three publicly available datasets to train and evaluate our system: DAGM2007~\cite{weimer2016}, KolektorSDD~\cite{Tabernik2019JIMkolek}, and KolektorSDD2~\cite{Bozic2021COMINDkolek2}.
Below is a brief overview of the three datasets:

\begin{itemize}
    \item \textbf{DAGM2007} contains grayscale images for 10 classes of artificially generated patterns, with around $8\%$ of them containing defects.
    The classes were designed to mirror real world problems, with 1150 images per class, and images of size $512\times512$.
    
    \item \textbf{KolektorSDD} is a small dataset of grayscale images collected from a real industrial environment.
    There are only 399 images, with around $8\%$ of them containing defects, and the standard size of images for the dataset being $512\times1408$.
    A sample image from KolektorSDD is shown in Figure~\ref{fig:sample_defect}.
    
    \item \textbf{KolektorSDD2} is a larger dataset of color images of size $230\times630$.
    There are $3335$ images with around $9\%$ of the images containing defects.
\end{itemize}

The raw data of the three datasets are stored in different directory hierarchies, and represent their annotations in varying formats.
Thus, we standardize our datasets to a common format, which simplifies our training and evaluation code later in the project.
This is the purpose of the \emph{Bonseyes Datatool} template, which provides utilities to create a conversion pipeline for raw data.
We created three datatools, one for each of our datasets, which all converged on a common format.
We represent a given dataset element with in the following format:

\begin{itemize}
    \item Path to image data.
    \item Compressed matrix representing the defect annotation.
    \item The classification: defective or non-defective.
\end{itemize}

The datatool represents all of this data in a standardized JSON format, with raw image data stored in a simple directory hierarchy.
Once the datatools have converted their respective datasets, we can then develop our training and evaluation pipeline.
To achieve this, we leverage the Bonseyes AI Asset system, which as discussed in Section~\ref{sec:core_tools} provides a set of utilities and packages for developing AI applications.
To ensure a consistent software environment we develop our solution in a Docker container provided by the AI Asset, adding any dependencies we require to the AI Asset's dependency file.


\subsection{Model architectures and training}
\label{subsec:models}

\begin{figure}
\centering
    \includegraphics[width=0.9\linewidth]{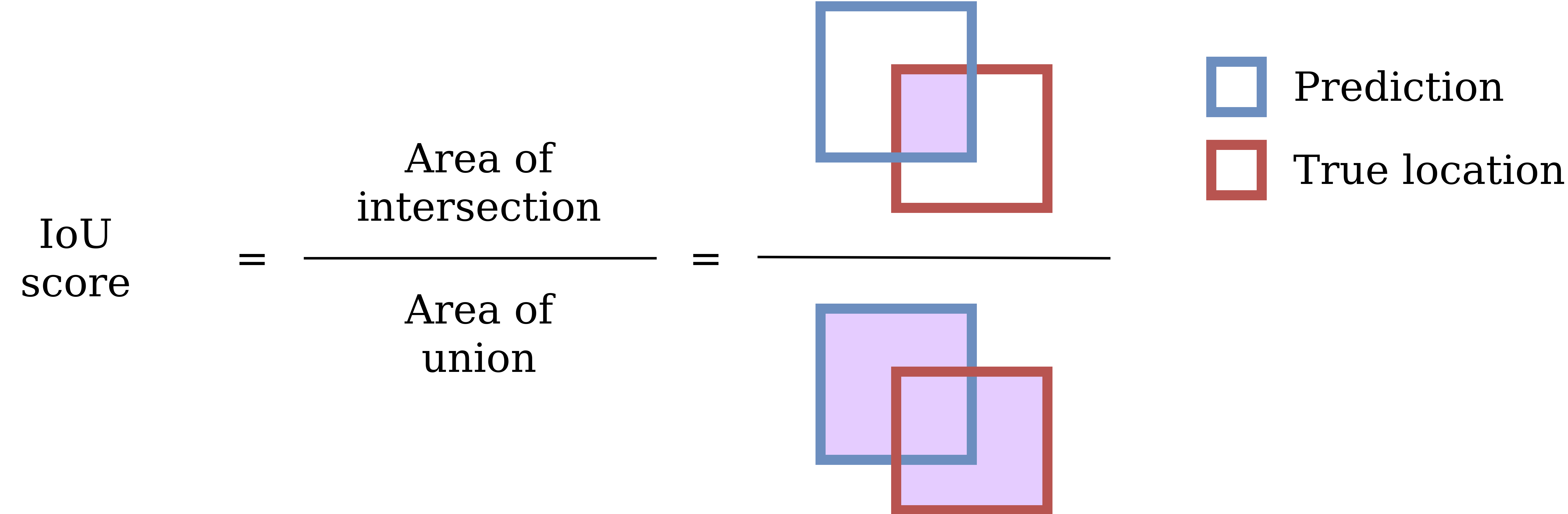}
\caption{Overview of the IoU-score which we optimize when training our DNN models.}
\label{fig:iou}
\end{figure}

Bonseyes AI Assets do not enforce any strict requirements on how models are developed, simply providing a template to follow.
Thus, for our models we leverage the SMP library~\cite{long2015Segmentation}, which provides model architectures for image segmentation.
We can formulate surface defect detection as an image segmentation problem by representing the annotations (e.g., the ones seen in Figure~\ref{fig:sample_label}) as a mask matrix of $1$s and $0$s for defective and non-defective pixels respectively.
Then when training we attempt to produce an output matrix which has maximum similarity with this matrix.
To measure this similarity, we use the common metric of intersection-over-union (IoU), as shown in Figure~\ref{fig:iou}, with an IoU-score of $0.0$ meaning that we predicted no defective pixels correctly, and $1.0$ meaning that we predicted every defective pixel correctly.
We leverage the fact that our three datasets are in the same format (as described in Section~\ref{subsec:datatools}) to simplify the definition of our models, since we only need to support one data format.

\begin{figure}
\centering
    \includegraphics[width=0.9\linewidth]{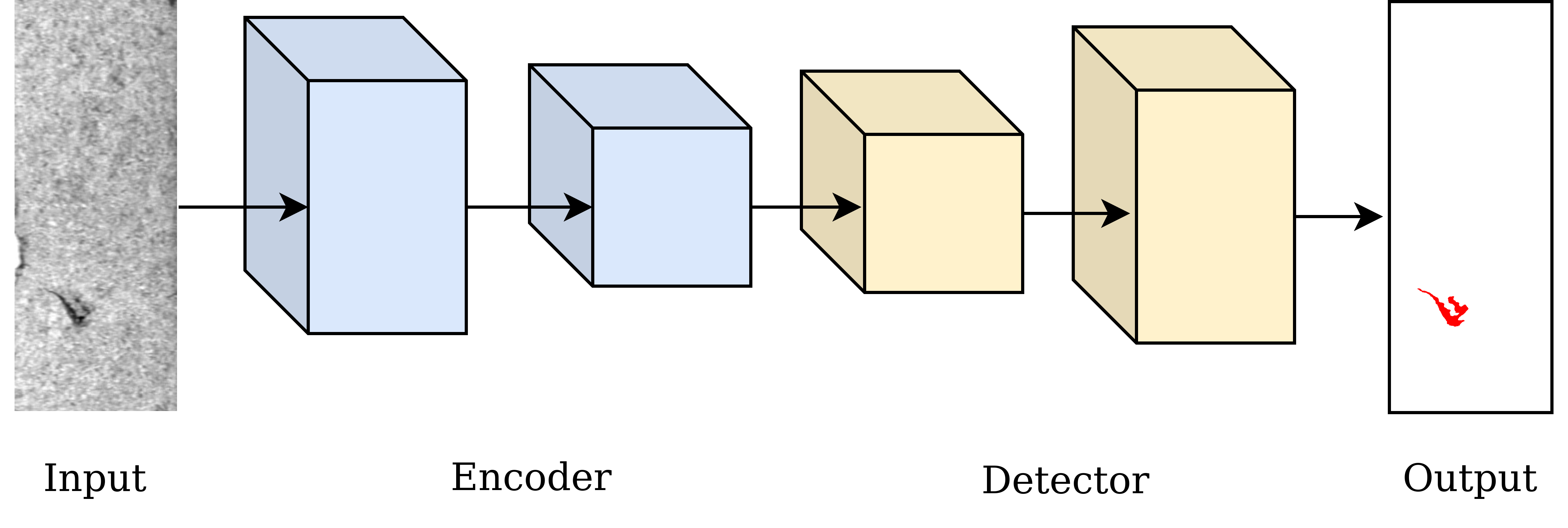}
\caption{
    Simplified example of an Encoder-Detector DNN architecture.
}
\label{fig:smp}
\end{figure}

Using the SMP library, DNN model architectures are defined with two main components: an \emph{encoder} model which processes raw image data,
and a model which takes the output of the encoder to produce the image segmentation, which we refer to as the \emph{detector}.
Thus in this paper we call this an encoder-detector architecture, as shown in Figure~\ref{fig:smp}.
An advantage of this architecture is that for the encoder we can leverage pretrained ImageNet~\cite{ILSVRC15_imagenet} models, such as ResNet50~\cite{he2016deep}, MobileNetV2~\cite{sandler2018mobilenetv2}, and EfficientNet~\cite{tan2019_effnet}.
This can significantly reduce our training costs and the amount of training data we require, since our models do not need to learn from scratch how to identify low-level image features (e.g., edges, corners, textures, etc).
The encoder model skips its final ImageNet classification layers, passing intermediate activations to the detector model.
This means that data passed to the detector is easier to process than raw image data.
SMP provides a number of state-of-the-art architectures we can choose for the detector, including Unet++~\cite{zhou2018a}, MAnet~\cite{fan2020}, LinkNet~\cite{chaurasia2017}, PAN~\cite{li2018}, and more.
In total, SMP can provide over $1000$ unique encoder-detector pairs, and in our evaluation in Section~\ref{sec:results} we train a subset ($62$ models) and report on our best performing models.

Using a higher level library such as SMP rather than building our own architecture from scratch, or using a single model implementation published alongside a research paper (e.g., \href{https://github.com/e-lab/LinkNet}{LinkNet}, \href{https://github.com/zengqunzhao/MA-Net}{MANet}) significantly increased our productivity, since we did not know ahead of time which architecture would provide the best performance, and having a tool such as SMP which allowed us to easily switch architectures meant we only had to integrate one codebase rather than several. 
In addition, for training our models we leveraged the PyTorch Lightning library~\cite{Falcon_PyTorch_Lightning_2019}, which further reduced the amount of boilerplate code we had to write for configuring the training procedures for our models.
For future work, PyTorch Lightning will also reduce the effort required to further compress our models using techniques such as pruning and quantization, which in our experience can be more difficult to do when only using the utilities provided by the base PyTorch library.

Our DNN models only provide a mask matrix, hence to provide a final classification we apply a post-processing step integrated into the Bonseyes AI Asset \emph{algorithm} class (which helps ensure that pre- and post-processing steps are performed consistently across frameworks).
The post-processing step takes a user-defined threshold (e.g., 1\%) and classifies the image as being defective if the proportion of pixels marked as defective by the model is greater than or equal to the threshold.
Thus, we can measure the quality of our solution using both classification accuracy and IoU score, however it is sufficient to train our models using the IoU score alone.

\subsection{Deployment, and complementary components}
\label{subsec:workflow}

Generally DNNs are trained on HPC servers featuring GPUs, however when they are deployed they may also run on more constrained edge devices such as IoT devices, smartphones, drones, VR headsets, etc.
When deploying on a new platform, tasks such as managing package dependencies and setting up the new environment can be very time consuming.
Fortunately, as discussed in Section~\ref{sec:core_tools}, Bonseyes AI Assets contain a CI/CD pipeline to ease this deployment.
Developers specify the versions of packages they are using for development purposes, and whenever they push a code commit the CI/CD pipeline builds the whole project for x86, Nvidia Jetson, and Raspberry Pi platforms.
Occasionally, a package version for the x86 platform may not be available for another platform.
In this case, developers will be sent an email about the issue and be able to specify a different package version for the platform, on in extreme cases a process to build the correct version of the package.
When the developer is ready to test deployment on their target device, they need only to download the latest version of the Docker image for their target platform.
Running CI/CD on a server with ISA emulation reduces the time required to setup a new platform, especially when steps like building packages can be more time consuming when compiling natively on constrained platforms such as the Raspberry Pi. 
This is especially valuable for machine learning researchers for whom testing on constrained edge devices may be considered merely supplementary results: not worth a large amount of effort, but whose inclusion will improve any evaluation they provide.

When running inference, deep learning frameworks such as PyTorch may not be the most optimal for running inference (since they are focused on training).
Thus, Bonseyes AI Assets also include support for ONNX Runtime and TensorRT inference, with support for more backends in development.

\section{Results}
\label{sec:results}

In this section, we discuss our results from training a range of models separately on our 3 datasets: DAGM2007, KolektorSDD, and KolektorSDD2.
In total we trained 62 models, and we include our tables of our 5 top performing models in terms of accuracy and inference time.
For all models, we evaluate them using a validation dataset, and for DAGM2007 and KolektorSDD2 we use the officially provided held-out test datasets.
For KolektorSDD, the dataset is too small for a held out test set (with only 52 examples of defects in the whole dataset).
Therefore, we evaluate it using the test set of KolektorSDD2, from which we expect to see a performance degradation due to the images having different characteristics.

We report the inference times in three settings:
\begin{enumerate*}
    \item PyTorch running on an NVidia A100 GPU,
    \item ONNX Runtime on a cloud-based Intel Broadwell series x86 CPU featuring 22 cores, and
    \item ONNX Runtime on the CPU of a Rapsberry Pi 4 Model B.
\end{enumerate*}
Note that we were unable to run models using EfficientNet-based encoders in the version of ONNX Runtime we tested due its lack of support for the models' ``Swish'' function.
Hence we represent the inference time for those models in this setting with a `-'.

Tables~\ref{tab:top_acc_dagm}, \ref{tab:top_acc_kol}, and \ref{tab:top_acc_kol2} show our top 5 models in terms of test set accuracy, as well as their inference time across our three settings.
Across every model we trained, our median test set classification accuracies were $99.8\%$, $89.0\%$, and $97.55\%$ for DAGM2007, KolektorSDD, and KolektorSDD2 respectively.
The lower accuracy for KolektorSDD is expected, given that 
\begin{enumerate*} 
\item our evaluation methodology tests on a completely different dataset (KolektorSDD2), and
\item how few examples are in the dataset relative to the other models.
\end{enumerate*}
We observe that across all of our DNN models, the mean validation set accuracy for our KolektorSDD models is $98.8\%$ (and was $99.7\%$ and $97.4\%$ for DAGM2007 and KolektorSDD2 respectively) which suggests that the models do learn well, however using KolektorSDD2 as a test set is unfair as the dataset is too different.
Comparing against other published approaches for our 3 datasets, as ranked by Papers with Code~\cite{paperswithcode2022c,paperswithcode2022a,paperswithcode2022b}, we observe that our best models get accuracies matching other highly ranked solutions.
We note that models using EfficientNetB4 as the encoder architecture appear disproportionately in the top-5 models in terms of accuracy for the 3 datasets, and there is no clear winner for detector architectures, suggesting that the choice of encoder architecture has the greatest influence on final accuracy.

Tables~\ref{tab:top_inf_dagm}, \ref{tab:top_inf_kol}, and \ref{tab:top_inf_kol2} show our top 5 models in terms of inference time on the NVidia A100 GPU, along with their accuracies, and inference time on other devices.
On the A100, our models vary in inference time between $5.9\mathrm{ms}$ and $34.0\mathrm{ms}$ running on an NVidia A100 with PyTorch.
We note that models with MobileNetV2 and ResNet34 as their encoder architectures are the only models that are in the top 5 in terms of inference time on the A100 across our three datasets, suggesting that as well as accuracy the encoder is the most important feature to consider for inference time.
We note that our fastest models see accuracy penalties when compared to their counterparts in Tables~\ref{tab:top_acc_dagm}, \ref{tab:top_acc_kol}, and \ref{tab:top_acc_kol2}.
However, several of our fast models also get high accuracies.
For example, in Table~\ref{tab:top_inf_dagm}, MobileNetV2-Unet (Rank 2) and MobileNetV2-Pan (Rank 5) get nearly perfect accuracy on the test set, and for Table~\ref{tab:top_inf_kol2} ResNet34-LinkNet (Rank 1) and MobileNetV2-Unet (Rank 2) get accuracies within $0.1\%$ of the fifth best performing model in Table~\ref{tab:top_acc_kol2}.
 
We observe that the relative inference times of models and their scaling does not necessarily stay the same between settings (i.e., PyTorch on the A100 GPU, ONNX Runtime the x86 CPU, ONNX Runtime on the Raspberry Pi 4).
For example in Table~\ref{tab:top_inf_dagm} MobileNetV2-Pan (Rank 5) is almost $3.3\times$ faster than ResNet34-Unet (Rank 4) on x86+ONNX Runtime, whereas the models have almost identical inference times on the A100 using PyTorch, with a similar discrepancy seen on the Raspberry Pi 4. 
This tells us that relative inference time performance is not necessarily consistent between frameworks and devices. 
In future work, we will explore in greater detail these performance trade-offs and variances, how to make the best choice of model for a given hardware platform, and investigate further across-stack DNN optimizations~\cite{turner2018a} such as grouped convolutions~\cite{gibson2020} and quantization.
Deep learning compilers such as TVM~\cite{tvm} and IREE~\cite{liuTinyIREEMLExecution2022} provide another dimension of DNN optimization, with approaches such as auto-tuning~\cite{chenLearningOptimizeTensor2018}, auto-scheduling~\cite{zhengAnsorGeneratingHighPerformance2020}, and related systems~\cite{gibsonReusingAutoSchedulesEfficient2022} potentially bringing further performance improvements.
Integration of these systems within an AI Asset could provide a more straightforward way to reap their benefits.

\begin{table*}[]
\centering
\footnotesize
\caption{Top 5 models ranked by accuracy for DAGM2007 along with their inference times.}
\label{tab:top_acc_dagm}
\begin{tabular}{|l|cr|rr|l|rrr|}
\hline
\multicolumn{1}{|c|}{\multirow{2}{*}{\textbf{Rank}}} &
  \multicolumn{2}{c|}{\textbf{Test}} &
  \multicolumn{2}{c|}{\textbf{Validation}} &
  \multicolumn{1}{c|}{\multirow{2}{*}{\textbf{Arch (Encoder-Detector)}}} &
  \multicolumn{3}{c|}{\textbf{Inf. time (ms)}} \\ \cline{2-5} \cline{7-9} 
\multicolumn{1}{|c|}{} &
  \multicolumn{1}{c|}{Acc (\%)} &
  \multicolumn{1}{c|}{IoU} &
  \multicolumn{1}{c|}{Acc (\%)} &
  \multicolumn{1}{c|}{IoU} &
  \multicolumn{1}{c|}{} &
  \multicolumn{1}{c|}{\textbf{A100}} &
  \multicolumn{1}{c|}{\textbf{x86}} &
  \multicolumn{1}{c|}{\textbf{RPi4}} \\ \hline
1 & \multicolumn{1}{c|}{100.0} & 0.941 & \multicolumn{1}{r|}{100}  & 0.976 & EfficientNetB4-LinkNet & \multicolumn{1}{r|}{18.9} & \multicolumn{1}{r|}{-}    & -    \\ \hline
2 & \multicolumn{1}{c|}{100.0} & 0.939 & \multicolumn{1}{r|}{100}  & 0.975 & EfficientNetB4-Pan     & \multicolumn{1}{r|}{21}   & \multicolumn{1}{r|}{-}    & -    \\ \hline
3 & \multicolumn{1}{c|}{100.0} & 0.937 & \multicolumn{1}{r|}{99.9} & 0.977 & EfficientNetB4-MANet   & \multicolumn{1}{r|}{23.5} & \multicolumn{1}{r|}{-}    & -    \\ \hline
4 & \multicolumn{1}{c|}{100.0} & 0.917 & \multicolumn{1}{r|}{100}  & 0.971 & MobileNetV2-Unet++      & \multicolumn{1}{r|}{8.2}  & \multicolumn{1}{r|}{116}  & 3351 \\ \hline
5 & \multicolumn{1}{c|}{100.0} & 0.914 & \multicolumn{1}{r|}{99.9} & 0.967 & MobileNetV2-Pan         & \multicolumn{1}{r|}{6.9}  & \multicolumn{1}{r|}{36.6} & 853  \\ \hline
\end{tabular}
\end{table*}

\begin{table*}[]
\centering
\footnotesize
\caption{Top 5 models ranked by accuracy for KolektorSDD along with their inference times.}
\label{tab:top_acc_kol}
\begin{tabular}{|l|rr|rr|l|rrr|}
\hline
\multicolumn{1}{|c|}{\multirow{2}{*}{\textbf{Rank}}} &
  \multicolumn{2}{c|}{\textbf{Test}} &
  \multicolumn{2}{c|}{\textbf{Validation}} &
  \multicolumn{1}{c|}{\multirow{2}{*}{\textbf{Arch (Encoder-Detector)}}} &
  \multicolumn{3}{c|}{\textbf{Inf. time (ms)}} \\ \cline{2-5} \cline{7-9} 
\multicolumn{1}{|c|}{} &
  \multicolumn{1}{c|}{Acc (\%)} &
  \multicolumn{1}{c|}{IoU} &
  \multicolumn{1}{c|}{Acc (\%)} &
  \multicolumn{1}{c|}{IoU} &
  \multicolumn{1}{c|}{} &
  \multicolumn{1}{c|}{\textbf{A100}} &
  \multicolumn{1}{c|}{\textbf{x86}} &
  \multicolumn{1}{c|}{\textbf{RPi4}} \\ \hline
1 &
  \multicolumn{1}{r|}{90.2} &
  0.892 &
  \multicolumn{1}{r|}{100} &
  0.931 &
  InceptionV4-Unet++ &
  \multicolumn{1}{r|}{34} &
  \multicolumn{1}{r|}{1427.6} &
  \multicolumn{1}{l|}{73372} \\ \hline
2 &
  \multicolumn{1}{r|}{90} &
  0.363 &
  \multicolumn{1}{r|}{98.8} &
  0.926 &
  MobileNetV2-Pan &
  \multicolumn{1}{r|}{8.6} &
  \multicolumn{1}{r|}{58.9} &
  2119 \\ \hline
3 &
  \multicolumn{1}{r|}{89.8} &
  0.79 &
  \multicolumn{1}{r|}{98.8} &
  0.93 &
  ResNet34-Unet++ &
  \multicolumn{1}{r|}{11.6} &
  \multicolumn{1}{r|}{608.0} &
  \multicolumn{1}{l|}{30656} \\ \hline
4 &
  \multicolumn{1}{r|}{89.7} &
  0.841 &
  \multicolumn{1}{r|}{100} &
  0.944 &
  EfficientNetB4-MANet &
  \multicolumn{1}{r|}{29} &
  \multicolumn{1}{r|}{-} &
  - \\ \hline
5 &
  \multicolumn{1}{r|}{89.6} &
  0.892 &
  \multicolumn{1}{r|}{98.8} &
  0.895 &
  EfficientNetB4-Unet &
  \multicolumn{1}{r|}{25.1} &
  \multicolumn{1}{r|}{-} &
  - \\ \hline
\end{tabular}
\end{table*}

\begin{table*}[]
\centering
\footnotesize
\caption{Top 5 models ranked by accuracy for KolektorSDD2 along with their inference times.}
\label{tab:top_acc_kol2}
\begin{tabular}{|l|rr|rr|l|rrr|}
\hline
\multicolumn{1}{|c|}{\multirow{2}{*}{\textbf{Rank}}} &
  \multicolumn{2}{c|}{\textbf{Test}} &
  \multicolumn{2}{c|}{\textbf{Validation}} &
  \multicolumn{1}{c|}{\multirow{2}{*}{\textbf{Arch (Encoder-Detector)}}} &
  \multicolumn{3}{c|}{\textbf{Inf. time (ms)}} \\ \cline{2-5} \cline{7-9} 
\multicolumn{1}{|c|}{} &
  \multicolumn{1}{c|}{Acc (\%)} &
  \multicolumn{1}{c|}{IoU} &
  \multicolumn{1}{c|}{Acc (\%)} &
  \multicolumn{1}{c|}{IoU} &
  \multicolumn{1}{c|}{} &
  \multicolumn{1}{c|}{\textbf{A100}} &
  \multicolumn{1}{c|}{\textbf{x86}} &
  \multicolumn{1}{c|}{\textbf{RPi4}} \\ \hline
1 &
  \multicolumn{1}{r|}{98.1} &
  0.842 &
  \multicolumn{1}{r|}{98.7} &
  0.857 &
  InceptionV4-Unet &
  \multicolumn{1}{r|}{23.7} &
  \multicolumn{1}{r|}{152.3} &
  \multicolumn{1}{l|}{5462} \\ \hline
2 &
  \multicolumn{1}{r|}{98} &
  0.952 &
  \multicolumn{1}{r|}{98.1} &
  0.947 &
  EfficientNetB4-Unet++ &
  \multicolumn{1}{r|}{25} &
  \multicolumn{1}{r|}{-} &
  - \\ \hline
3 &
  \multicolumn{1}{r|}{98} &
  0.948 &
  \multicolumn{1}{r|}{97.2} &
  0.944 &
  EfficientNetB4-Pan &
  \multicolumn{1}{r|}{21.8} &
  \multicolumn{1}{r|}{-} &
  - \\ \hline
4 &
  \multicolumn{1}{r|}{97.9} &
  0.882 &
  \multicolumn{1}{r|}{97.9} &
  0.879 &
  EfficientNetB4-Unet &
  \multicolumn{1}{r|}{20.2} &
  \multicolumn{1}{r|}{-} &
  - \\ \hline
5 &
  \multicolumn{1}{r|}{97.8} &
  0.94 &
  \multicolumn{1}{r|}{97.9} &
  0.939 &
  EfficientNetB4-LinkNet &
  \multicolumn{1}{r|}{21} &
  \multicolumn{1}{r|}{-} &
  - \\ \hline
\end{tabular}
\end{table*}

\begin{table*}[]
\centering
\caption{Top 5 models ranked by inference time on the NVidia A100 for DAGM2007 along with their accuracies, and inference times on other platforms.}
\label{tab:top_inf_dagm}
\begin{tabular}{|l|rr|rr|l|rrr|}
\hline
\multicolumn{1}{|c|}{\multirow{2}{*}{\textbf{Rank}}} &
  \multicolumn{2}{c|}{\textbf{Test}} &
  \multicolumn{2}{c|}{\textbf{Validation}} &
  \multicolumn{1}{c|}{\multirow{2}{*}{\textbf{Arch (Encoder-Detector)}}} &
  \multicolumn{3}{c|}{\textbf{Inf. time (ms)}} \\ \cline{2-5} \cline{7-9} 
\multicolumn{1}{|c|}{} &
  \multicolumn{1}{c|}{Acc (\%)} &
  \multicolumn{1}{c|}{IoU} &
  \multicolumn{1}{c|}{Acc (\%)} &
  \multicolumn{1}{c|}{IoU} &
  \multicolumn{1}{c|}{} &
  \multicolumn{1}{c|}{\textbf{A100}} &
  \multicolumn{1}{c|}{\textbf{x86}} &
  \multicolumn{1}{c|}{\textbf{RPi4}} \\ \hline
1 & \multicolumn{1}{r|}{94.1} & 0.848 & \multicolumn{1}{r|}{97.4} & 0.941 & MobileNetV2-LinkNet & \multicolumn{1}{r|}{5.9} & \multicolumn{1}{r|}{43.3}  & 766  \\ \hline
2 & \multicolumn{1}{r|}{99.9} & 0.908 & \multicolumn{1}{r|}{99.9} & 0.968 & MobileNetV2-Unet    & \multicolumn{1}{r|}{6.4} & \multicolumn{1}{r|}{73.4}  & 2514 \\ \hline
3 & \multicolumn{1}{r|}{94.9} & 0.856 & \multicolumn{1}{r|}{97.2} & 0.941 & ResNet34-Pan        & \multicolumn{1}{r|}{6.6} & \multicolumn{1}{r|}{102.1} & 4459 \\ \hline
4 & \multicolumn{1}{r|}{94.5} & 0.844 & \multicolumn{1}{r|}{97.1} & 0.939 & ResNet34-Unet       & \multicolumn{1}{r|}{6.8} & \multicolumn{1}{r|}{120.2} & 4948 \\ \hline
5 & \multicolumn{1}{r|}{100}  & 0.914 & \multicolumn{1}{r|}{99.9} & 0.967 & MobileNetV2-Pan     & \multicolumn{1}{r|}{6.9} & \multicolumn{1}{r|}{36.6}  & 852  \\ \hline
\end{tabular}
\end{table*}

\begin{table*}[]
\centering
\caption{Top 5 models ranked by inference time on the NVidia A100 for KolektorSDD along with their accuracies, and inference times on other platforms.}
\label{tab:top_inf_kol}
\begin{tabular}{|l|rr|rr|l|rrr|}
\hline
\multicolumn{1}{|c|}{\multirow{2}{*}{\textbf{Rank}}} &
  \multicolumn{2}{c|}{\textbf{Test}} &
  \multicolumn{2}{c|}{\textbf{Validation}} &
  \multicolumn{1}{c|}{\multirow{2}{*}{\textbf{Arch (Encoder-Detector)}}} &
  \multicolumn{3}{c|}{\textbf{Inf. time (ms)}} \\ \cline{2-5} \cline{7-9} 
\multicolumn{1}{|c|}{} &
  \multicolumn{1}{c|}{Acc (\%)} &
  \multicolumn{1}{c|}{IoU} &
  \multicolumn{1}{c|}{Acc (\%)} &
  \multicolumn{1}{c|}{IoU} &
  \multicolumn{1}{c|}{} &
  \multicolumn{1}{c|}{\textbf{A100}} &
  \multicolumn{1}{c|}{\textbf{x86}} &
  \multicolumn{1}{c|}{\textbf{RPi4}} \\ \hline
1 &
  \multicolumn{1}{r|}{89} &
  0.696 &
  \multicolumn{1}{r|}{100} &
  0.899 &
  MobileNetV2-LinkNet &
  \multicolumn{1}{r|}{7.9} &
  \multicolumn{1}{r|}{80.2} &
  1914 \\ \hline
2 &
  \multicolumn{1}{r|}{89} &
  0.154 &
  \multicolumn{1}{r|}{96.2} &
  0.918 &
  ResNet34-LinkNet &
  \multicolumn{1}{r|}{8} &
  \multicolumn{1}{r|}{258.1} &
  8757 \\ \hline
3 &
  \multicolumn{1}{r|}{64.4} &
  0.002 &
  \multicolumn{1}{r|}{98.8} &
  0.924 &
  ResNet34-Unet &
  \multicolumn{1}{r|}{8} &
  \multicolumn{1}{r|}{256.8} &
  \multicolumn{1}{l|}{12431} \\ \hline
4 &
  \multicolumn{1}{r|}{89} &
  0.89 &
  \multicolumn{1}{r|}{98.8} &
  0.903 &
  ResNet34-DeepLabV3 &
  \multicolumn{1}{r|}{8.1} &
  \multicolumn{1}{r|}{642.0} &
  40772 \\ \hline
5 &
  \multicolumn{1}{r|}{89} &
  0.89 &
  \multicolumn{1}{r|}{98.8} &
  0.925 &
  ResNet34-Pan &
  \multicolumn{1}{r|}{8.2} &
  \multicolumn{1}{r|}{227.1} &
  12221 \\ \hline
\end{tabular}
\end{table*}

\begin{table*}[]
\centering
\caption{Top 5 models ranked by inference time on the NVidia A100 for KolektorSDD2 along with their accuracies, and inference times on other platforms.}
\label{tab:top_inf_kol2}
\begin{tabular}{|l|rr|rr|l|rrr|}
\hline
\multicolumn{1}{|c|}{\multirow{2}{*}{\textbf{Rank}}} &
  \multicolumn{2}{c|}{\textbf{Test}} &
  \multicolumn{2}{c|}{\textbf{Validation}} &
  \multicolumn{1}{c|}{\multirow{2}{*}{\textbf{Arch (Encoder-Detector)}}} &
  \multicolumn{3}{c|}{\textbf{Inf. time (ms)}} \\ \cline{2-5} \cline{7-9} 
\multicolumn{1}{|c|}{} &
  \multicolumn{1}{c|}{Acc (\%)} &
  \multicolumn{1}{c|}{IoU} &
  \multicolumn{1}{c|}{Acc (\%)} &
  \multicolumn{1}{c|}{IoU} &
  \multicolumn{1}{c|}{} &
  \multicolumn{1}{c|}{\textbf{A100}} &
  \multicolumn{1}{c|}{\textbf{x86}} &
  \multicolumn{1}{c|}{\textbf{RPi4}} \\ \hline
1 & \multicolumn{1}{r|}{97.7} & 0.819 & \multicolumn{1}{r|}{97.4} & 0.817 & ResNet34-LinkNet      & \multicolumn{1}{r|}{6}   & \multicolumn{1}{r|}{71.4}  & 1908 \\ \hline
2 & \multicolumn{1}{r|}{97.7} & 0.94  & \multicolumn{1}{r|}{97.6} & 0.937 & MobileNetV2-Unet      & \multicolumn{1}{r|}{6.1} & \multicolumn{1}{r|}{60.0}  & 1539 \\ \hline
3 & \multicolumn{1}{r|}{95.2} & 0.887 & \multicolumn{1}{r|}{93.1} & 0.863 & MobileNetV2-LinkNet   & \multicolumn{1}{r|}{6.2} & \multicolumn{1}{r|}{32.1}  & 441  \\ \hline
4 & \multicolumn{1}{r|}{96.1} & 0.924 & \multicolumn{1}{r|}{95.5} & 0.907 & ResNet34-Unet         & \multicolumn{1}{r|}{6.4} & \multicolumn{1}{r|}{76.7}  & 2927 \\ \hline
5 & \multicolumn{1}{r|}{97.4} & 0.937 & \multicolumn{1}{r|}{97.9} & 0.939 & MobileNetV2-DeepLabV3 & \multicolumn{1}{r|}{6.6} & \multicolumn{1}{r|}{106.8} & 4552 \\ \hline
\end{tabular}
\end{table*}

\section{Conclusion}

In conclusion, there are a wide range of tools available to increase the productivity of both deep learning engineers and researchers.
Our paper highlighted that features such a continuous integration and deployment, which are rarely a priority for researchers, can bring a number of benefits and require little effort to set up if researchers embrace predefined workflows such as the open source tools provided by the Bonseyes Marketplace~\cite{llewellynn2017}.
In addition, there are emerging higher level deep learning libraries (such as SMP~\cite{Yakubovskiy:2019}) that can improve productivity for specific domains, that should be exploited where possible.
We discussed our experience using the latest tools to produce a solution for the problem of industrial defect detection, presenting results on an HPC server and a Raspberry Pi 4.
For future work, we will seek to continue to use these tools and principles to improve the quality of our own research artifacts, and explore the utilities provided by both PyTorch Lightning and Bonseyes AI Assets for model compression such as pruning and quantization.

\section*{Acknowledgments}

This project has received funding from the European Union’s Horizon 2020 research and innovation programme within the framework of the BonsAPPs Project funded under grant agreement No 101015848.
We thank Miguel De Prado for his insightful review and suggestions.

\balance
\bibliography{references}
\bibliographystyle{IEEEtran}

\end{document}